\documentclass{article}

\usepackage[preprint,nonatbib]{neurips_2026}

\usepackage[utf8]{inputenc} 
\usepackage[T1]{fontenc}    
\usepackage{hyperref}       
\usepackage{url}            
\usepackage{booktabs}       
\usepackage{amsfonts}       
\usepackage{nicefrac}       
\usepackage{microtype}      
\usepackage{xcolor}         
\usepackage{graphicx}       
\usepackage{multirow}       
\usepackage[numbers]{natbib} 
\usepackage{wrapfig}        

\usepackage{algorithm}
\usepackage{algorithmic}

\usepackage[frozencache=true,cachedir=.]{minted} 

\usepackage{newfloat}
\usepackage{listings}
\usepackage{caption}        
\DeclareCaptionStyle{ruled}{labelfont=normalfont,labelsep=colon,strut=off}
\floatstyle{ruled}
\newfloat{listing}{tb}{lst}{}
\floatname{listing}{Listing}

\newcommand{\Apps}{\mathcal{A}}
\newcommand{\app}[1]{a_{#1}}
\newcommand{\API}[1]{\mathrm{API}(#1)}
\newcommand{\state}{s}
\newcommand{\initstate}{s_0}
\newcommand{\goalstate}{s_g}

\newcommand{\task}{\tau}
\newcommand{\instruction}{I}
\newcommand{\policy}{\pi}
\newcommand{\policyi}[1]{\pi_{#1}}

\newcommand{\taskset}[1]{\{\tau_1, \tau_2, \dots, \tau_{#1}\}}

\newcommand{\numtasks}{k}

\newcommand{\threshold}{\tau}
\newcommand{\topk}{k}

\newcommand{\methodname}{HCL-GP\ }
\newcommand{\baselinemethodname}{GP\ }

\title{Learning and Reusing Policy Decompositions for \\ Hierarchical Generalized Planning with LLM Agents}

\author {
\hspace{-0.5cm}
Shirin Sohrabi \;\; Haritha Ananthakrishnan \;\; Harsha Kokel \;\; Kavitha Srinivas \;\; Michael Katz\\
\hspace{-0.5cm}
IBM\\
\texttt{ssohrab@us.ibm.com}\\ \texttt{\{hananthakris,Harsha.Kokel,kavitha.srinivas,Michael.Katz1\}@ibm.com}  
}

\begin{document}

\maketitle

\begin{abstract}

We present a dynamic policy-learning approach that combines generalized planning and hierarchical task decomposition for LLM-based agents. Our method, Hierarchical Component Learning for Generalized Policies (\methodname{}), learns parameterized policies that generalize across task instances and automatically extracts reusable components from successful executions, organizing them into a component library for compositional policy generation. We address three challenges: (1) learning components through automated decomposition, (2) generalizing components to maximize reuse, and (3) efficient retrieval via semantic search. 

Evaluated on the AppWorld benchmark, our approach achieves 98.2\% accuracy on normal tasks and 97.8\% on challenge tasks with unseen applications, improving 15.8 points over static synthesis on challenging scenarios. For open-source models, dynamic reuse enables 62.5\% success versus near-zero without reuse. This demonstrates that classical planning concepts can be effectively integrated with LLM agents for improved accuracy and efficiency.

\end{abstract}

\section{Introduction}

Interactive coding agents solve realistic tasks by generating executable code that queries APIs, manipulates state, and iteratively repairs failures. Yet many systems treat each task in isolation, repeatedly rediscovering solution patterns and failing to accumulate reusable procedural knowledge.

We study dynamic policy learning in structured sequential decision-making settings where environments contain multiple domains, each consisting of related task instances that share latent structure but differ in parameters. The objective is to produce generalized, reusable policies that instantiate across instances and adapt efficiently to new domains.

Our approach draws on two AI planning paradigms: generalized planning \cite{bonet-et-al-ijcai2017,hodel-2024,silver-et-al-aaai2024,stein-et-al-icaps2026} produces parameterized policies that solve classes of problems by abstracting over instance-specific details and Hierarchical Task Network (HTN) planning \cite{erol-et-al-aaai1994,hogg-et-al-aaai2008,georgievski-aiello-aij2015} decomposes complex tasks into reusable sub-tasks. We integrate these principles with LLM-based agents through a dynamic policy-learning framework.

Our novelty lies in realizing generalized and hierarchical planning in a model-free LLM-agent setting through execution-grounded learning of reusable executable policy components. We do so without symbolic domain models, hand-coded decomposition structures, or explicit transition models. Instead, reusable policy structure is induced dynamically from successful executions, validated through interaction, generalized across domains, and used compositionally for future policy synthesis.

We instantiate this framework on AppWorld \cite{trivedi-et-al-acl-findings2024}, a benchmark where agents interact with multiple applications through APIs. Each scenario consists of multiple related tasks sharing structure but varying in parameters, making AppWorld a natural testbed for generalized and compositional policies.

Our contributions are: (1) a dynamic policy-learning framework integrating generalized planning and hierarchical decomposition for reusable, parameterized policies; (2) methods for automatically extracting, validating, generalizing, and reusing executable policy components from successful executions; and (3) an empirical evaluation on AppWorld demonstrating improved efficiency and favorable cost-accuracy trade-offs, particularly under challenging cross-domain transfer, showing that classical planning concepts remain effective tools for structuring LLM-based agents.
\section{Related Work}

Our work is grounded in generalized planning and hierarchical task decomposition. Generalized planning synthesizes policies that solve classes of problems by abstracting instance-specific details \cite{jimenez-et-al-ker2019}. Recent work uses LLMs to generate such policies in code \cite{silver-et-al-aaai2024,stein-et-al-icaps2026}. We learn parameterized policies and reusable components that generalize dynamically across instances. Unlike traditional approaches, we do not assume symbolic domain representations; generalization emerges from execution feedback and iterative synthesis.

Hierarchical Task Network (HTN) planning decomposes tasks into reusable subtasks \cite{erol-et-al-aaai1994,nau-et-al-jair2003,ghallab-et-al-2004}. HTN learning aims to automatically identify reusable substructures \cite{hogg-et-al-aaai2008,georgievski-aiello-aij2015}. We adopt this insight but learn hierarchical structure implicitly from LLM-generated executions rather than symbolic plans, and discover components dynamically rather than enforcing fixed task networks.

Learning reusable skills has been studied in learning from demonstrations \cite{argall-et-al-ras2009} and hierarchical RL
\cite{sutton-et-al-aij1999,hutsebaut-buysse-et-al-make2022}. These methods typically operate in low-level control settings with fixed state-action spaces. We differ by learning reusable skills as executable code fragments that 
implement policy logic.

Our work lies at the intersection of classical planning and modern LLM-based agents, demonstrating how generalized planning and hierarchical decomposition can be realized dynamically without symbolic domain models by extracting, evaluating, and reusing executable policy components learned through interaction.

\section{Problem Formulation}

We consider a structured sequential decision-making setting organized around a \emph{meta-domain} $\mathcal{M}$ that defines a shared action space and deterministic execution semantics.

Within a meta-domain, we distinguish between \emph{domains} and \emph{task instances}. A domain $\mathcal{D}$ is a collection of related task instances that share latent procedural structure but differ in parameters such as entities, values, or initial conditions. A task instance $\task = (\instruction, \initstate, \goalstate)$ consists of:
$\instruction$: a natural-language instruction describing the desired behavior,
$\initstate$: an initial state, and
$\goalstate$: a goal specification defining correctness.

Following generalized planning terminology, we distinguish between policies and plans. A \textbf{policy} $\policy$ is a parameterized program with signature $\sigma(\policy)$ that maps task parameters to executable plans. To solve a task instance $\task$, the system must: (1) extract parameters $\theta$ from $\task$ that match $\sigma(\policy)$, (2) generate a \textbf{plan} $\pi(\theta)$ by instantiating $\policy$ with $\theta$, and (3) submit the plan along with $\initstate$ and $\goalstate$ to a \textbf{validator}. The validator executes $\pi(\theta)$ from $\initstate$ and checks whether the resulting state satisfies $\goalstate$. The validator provides feedback indicating success or failure, and may include error messages or execution traces to guide debugging.

The key structural property of this setting is that task instances within a domain share underlying workflows or procedural patterns, differing primarily in parameter values. This structure enables learning parameterized policies that generalize across instances within a domain, and extracting reusable components that transfer across domains within the same meta-domain.

\vspace{-1ex}
\paragraph{Objective:}
Given domains $\{\mathcal{D}_1, \dots, \mathcal{D}_n\}$ within meta-domain $\mathcal{M}$, we synthesize parameterized policies that solve task instances efficiently and compositionally.
For each domain $\mathcal{D}_i$, we seek a parameterized policy $\policyi{i}$ with signature $\sigma(\policyi{i})$ that solves all instances in $\mathcal{D}_i$. The policy captures shared procedural structure, while the signature defines varying parameters. Instantiating $\policyi{i}$ with task-specific parameters $\theta_j$ produces a plan satisfying task requirements.

Beyond solving individual domains, we leverage experience from solved domains to solve new domains efficiently by identifying and reusing recurring procedural patterns. The challenge is learning policies that generalize across task instances within domains and compose reusable knowledge across domains, enabling efficient adaptation without complete policy resynthesis.

\paragraph{AppWorld as a Meta-Domain Instantiation:}
We instantiate this general formulation on the AppWorld benchmark \cite{trivedi-et-al-acl-findings2024}, which provides a concrete realization of the meta-domain structure.

\textbf{Meta-domain:} The AppWorld benchmark defines the meta-domain $\mathcal{M}$. The environment consists of a set of applications $\Apps = \{\app{1}, \app{2}, \dots, \app{m}\}$ (e.g., Gmail, Spotify, Phone), where each application $\app{i}$ exposes a collection of APIs denoted by $\API{\app{i}}$. The global world state $\state$ is represented as a database encoding the state of all applications. The agent cannot access or modify $\state$ directly; all interaction occurs exclusively through API calls, which deterministically update the world state.

\textbf{Domains:} In AppWorld, each \emph{scenario} corresponds to a domain $\mathcal{D}$. A scenario is a set of tasks $\taskset{\numtasks}$.
Tasks within a scenario invoke the same 
APIs, and differ primarily in parameter values such as entities, amounts, dates, or labels. For example, a payment scenario may contain 
tasks that differ only in recipients, amounts, or notes, while sharing the same high-level workflow.

\textbf{Task instances:} 
Each task $\task = (\instruction, \initstate, \goalstate)$ requires the system to generate a parameterized policy (a Python function) and instantiate it with task-specific parameters to produce an executable plan. The plan consists of a sequence of API calls and glue code (such as processing responses, manipulating results, etc). Task correctness is evaluated by executing the plan in the AppWorld simulator and comparing the resulting state to the expected effects encoded in $\goalstate$ through per-task test cases.

This instantiation makes AppWorld a natural testbed for evaluating generalized and compositional policies: scenarios provide natural domain boundaries, tasks within scenarios share procedural structure, and the benchmark includes multiple scenarios that exhibit recurring patterns across different applications and workflows.

\section{Approach}

We present a meta-domain-agnostic architecture for dynamic policy learning that is fully transferable across problem settings. The approach requires only that policies are executable programs, a validator checks task correctness and provides an interpretable feedback. The architecture does not depend on specific action spaces, state representations, or symbolic domain models, but relies on agents capable of semantic understanding and code generation from natural language, making it adaptable across meta-domains without modification.

\begin{figure*}[t]
\centering
\includegraphics[width=0.8\textwidth]{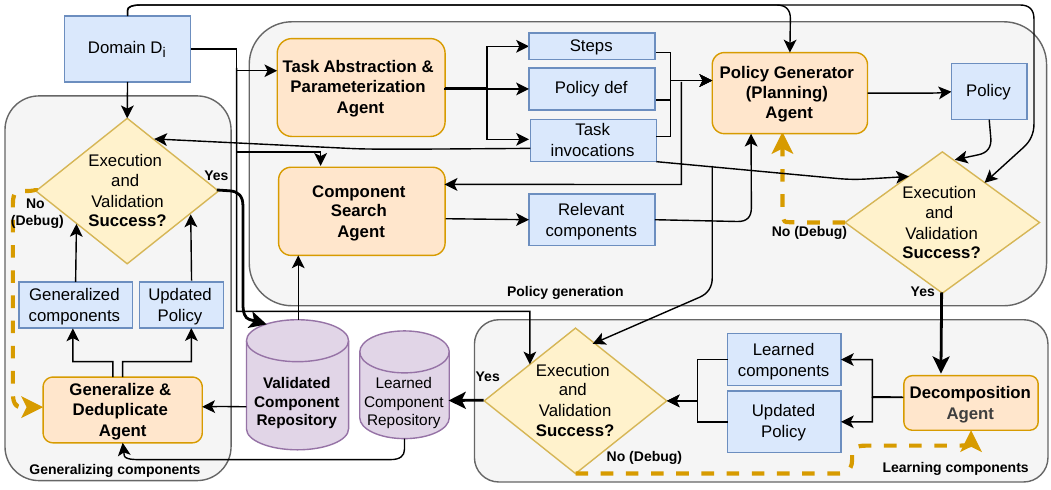}
\caption{Dynamic policy-learning architecture with three sections: (1) \textbf{Policy generation} synthesizes policies using task abstraction, component search, and iterative validation; (2) \textbf{Learning components} extracts reusable components from successful policies; (3) \textbf{Generalizing components} consolidates and validates components before storing them in the repository for subsequent use.}
\label{fig:architecture}
\vspace{-1ex}
\end{figure*}

Figure~\ref{fig:architecture} provides an overview of the architecture and the interactions among its components. Agents are marked by orange boxes in the figure. The architecture consists of three main sections that operate on a given domain: (1) \textbf{Policy generation} synthesizes a parameterized policy through task abstraction, component retrieval, and iterative validation with debugging; (2) \textbf{Learning components} extracts reusable components from the successful policy via decomposition and validates them; (3) \textbf{Generalizing components} consolidates learned components with previously validated ones through clustering and generalization, with its own validation loop to ensure correctness is preserved. The validated component repository is used for component retrieval in subsequent domains.

\paragraph{Policy Generation.}
Policy synthesis proceeds through the flow shown in the policy generation section of Figure~\ref{fig:architecture}. Given a domain $\mathcal{D}_i$, the \textbf{Task Abstraction and Parameterization Agent} induces a high-level representation capturing shared structure across task instances. This abstraction produces three key artifacts: (1) high-level steps describing the workflow, (2) a policy signature $\sigma(\policy)$ specifying the parameters and interface of the domain-level policy, and (3) concrete parameter bindings $\{\theta_1, \dots, \theta_k\}$ for each task instance. This abstraction enables generalization across task instances within the domain and guides downstream reasoning. The abstraction process is meta-domain-agnostic: it operates purely on natural language descriptions and does not require knowledge of the specific action space or state representation.

To support compositional policy generation, the \textbf{Component Search Agent} retrieves relevant reusable components from the validated component repository. Specifically, all validated components are embedded and indexed, and only the top‑$k$ most relevant components are retrieved for a given domain based on semantic similarity to the domain abstraction. The retrieved components, restricted to their signatures and usage information, are provided as context to the \textbf{Policy Generator (Planning) Agent}, which is then tasked with generating a parameterized policy $\policy$ for the domain in the appropriate executable representation (e.g., Python functions for AppWorld). The Policy Generator receives four inputs: the high-level steps, the policy signature $\sigma(\policy)$, the parameter bindings, and the relevant components from the Component Search Agent.

The synthesized policy is instantiated with task-specific parameters to produce plans, which are submitted to the validator.
If validation fails, error messages and execution traces are returned to the Policy Generator Agent, which 
revises the policy. This validation-debug cycle continues until successful plans are obtained for all task instances or a fixed 
budget is exhausted. This feedback loop is meta-domain-agnostic: it requires only that the validator provides interpretable error signals.

\paragraph{Learning Components.}
When a domain-level policy succeeds, the \textbf{Decomposition Agent} analyzes the policy and extracts reusable components corresponding to coherent fragments of executable logic that may be applicable to other domains. The Decomposition Agent operates through code analysis, identifying procedural patterns and abstractions without requiring domain-specific heuristics. The agent produces two outputs: (1) learned components representing the extracted reusable fragments, and (2) an updated policy that references these components. Both outputs are validated by instantiating the updated policy with the original parameter bindings and submitting the resulting plans to the validator across all task instances within the domain. Upon successful validation, the learned components are stored in the learned component repository, while the updated policy replaces the original. At this stage, components are accumulated but have not yet been generalized or deduplicated.

\paragraph{Generalizing Components.}
The system evaluates and generalizes components from the learned repository together with previously validated components (Figure~\ref{fig:architecture}). This stage achieves transfer across domains within the meta-domain. The generalization process can be triggered after processing a fixed number of domains, when the repository reaches a size threshold, or periodically.
Candidate components are clustered based on functional similarity using semantic embeddings of executable code. Clustering identifies groups of related components and separates distinct ones, improving efficiency and quality of generalization.

The \textbf{Generalize and Deduplicate Agent} consolidates redundant or overly specific components within each cluster into general abstractions. For example, \texttt{login\_to\_phone} and \texttt{login\_to\_venmo} might merge into \texttt{login\_to\_app}. The agent produces: (1) generalized components and (2) updated policies using these components.
Both outputs are validated before acceptance. Updated policies are instantiated with original parameters and submitted to the validator with corresponding task instances, ensuring generalization preserves correctness. If validation fails, the agent enters a debugging loop until components pass all tests or a budget is exhausted. Successful components are stored in the validated repository for retrieval in subsequent domains.
This stage consolidates reusable knowledge and induces hierarchical structure over executable components. The validated repository provides a growing library supporting efficient compositional policy generation.

All agents operate through semantic understanding and code generation, making the approach representation-agnostic. The architecture adapts to different meta-domains by learning appropriate executable representations through validator interaction, without requiring manual domain models. This enables application to web automation, database management, API orchestration, or other structured settings with executable policies and interpretable feedback.

\section{Experimental Evaluation}

We aim to answer three key research questions:

\noindent\textbf{Efficiency of reuse:} Does dynamic learning and reuse of policy components reduce the computational effort required to solve new task instances compared to synthesizing policies from scratch?

\noindent\textbf{Cost-accuracy trade-offs:} How does the proposed approach balance inference cost against task success, particularly in resource-constrained settings?

\noindent\textbf{Generalization across domains:} Can components learned from one set of scenarios transfer effectively to previously unseen scenarios, including those involving different applications?

To address these questions, we evaluate our approach on AppWorld, which provides a structured environment for multi-step decision-making with shared procedural patterns across scenarios. We measure performance not only in terms of final task success, but also as a function of debugging iterations and cumulative inference cost, providing insight into efficiency gains from dynamic reuse.

\subsection{AppWorld Implementation Details}

\paragraph{Policy Generation.} Policy synthesis is guided by semantic retrieval over both APIs and reusable components. All validated components are embedded and indexed using the \texttt{BAAI/bge-large-en-v1.5} (BAAI) embedding model. For each domain (scenario in AppWorld), the Component Search Agent retrieves the top‑$\topk$ most relevant reusable components based on cosine similarity to the domain abstraction. We set $\topk=20$ for component retrieval. The Policy Generator Agent also retrieves the top‑$\topk=20$ most relevant APIs from the AppWorld API documentation using the same embedding model and similarity metric.

\paragraph{Learning Components.} The Decomposition Agent analyzes successful domain-level policies to extract reusable components. In the AppWorld instantiation, the agent identifies Python functions that implement coherent subtasks (e.g., authentication workflows, data retrieval patterns, or transaction sequences). The agent produces both the extracted components and an updated policy that references them. Validation ensures that the updated policy produces identical behavior to the original across all task instances in the domain.

\paragraph{Generalizing Components.} Component clustering uses the BAAI embedding model to compute pairwise cosine similarity over all candidate components from the learned repository. We apply threshold-based clustering with $\threshold = 0.85$, followed by a greedy clustering procedure to obtain initial clusters. The Generalize and Deduplicate Agent processes each cluster to consolidate redundant or overly specific components. For example, in AppWorld, the agent may merge domain-specific authentication functions like \texttt{login\_to\_phone} and \texttt{login\_to\_venmo} into a single parameterized \texttt{login\_to\_app} component. The agent can also subsume entire clusters by producing higher-level abstractions. Both the generalized components and updated policies are validated against all original task instances before being accepted into the validated component repository.

\paragraph{Execution and Validation.} Each generated policy is executed within the AppWorld environment, which provides a sandboxed Python runtime with access to simulated applications and APIs. The execution environment captures all API calls, return values, and any runtime errors or exceptions. Validation is performed by comparing the policy's execution trace and final state against the ground-truth goal conditions specified for each instance. The environment then reports success or failure. 

For domain-level policies, we require success across all task instances within the scenario. When learning or generalizing components, the system validates success of the updated policies (which reference the new components). 
We set the debugging budget to three iterations per validation cycle, allowing the Policy Generator Agent to iteratively refine policies based on execution feedback (error messages, API responses, or partial results) before a policy is deemed unsuccessful.

\paragraph{Prompts and Implementation.} The prompts used for LLM calls to the Task Abstraction and Parameterization Agent, Policy Generator Agent, Decomposition Agent, and Generalize and Deduplicate Agent are provided in the supplementary material. 

\subsection{Experimental Setup}

\paragraph{Benchmark.} We use the AppWorld benchmark \cite{trivedi-et-al-acl-findings2024}, which consists of scenarios organized into training, normal test, and challenge test sets. Each scenario contains multiple related tasks that share underlying workflows but differ in parameters. The challenge test set includes applications (Gmail, Amazon) that do not appear in the training set, enabling evaluation of cross-application transfer.

AppWorld defines two primary evaluation metrics. In \textbf{Task Goal Completion (TGC)}, success is measured independently per task, providing a granular view of performance on individual task instances. In \textbf{Scenario Goal Completion (SGC)}, a scenario is counted as successful only if all tasks within the scenario are solved. This metric directly evaluates the ability to synthesize policies that generalize consistently across task instances within a scenario.
We report both metrics where applicable, and analyze performance as a function of debugging iterations and cumulative inference cost to assess efficiency alongside accuracy.

\paragraph{Our Method.} We evaluate our dynamic policy-learning architecture, which includes: Task Abstraction and Parameterization Agent, Component Search Agent, Policy Generator Agent, Decomposition Agent, and Generalize and Deduplicate Agent. The model-agnostic architecture operates through semantic understanding and code generation from natural language descriptions.

To obtain the initial seed for the validated component repository, we run the full architecture on the 30 scenarios in the AppWorld training set, starting from an empty component repository. For each training scenario, we perform policy generation and component learning as described in the previous section. After processing all training scenarios, the learned component repository contains components extracted from successful policies across the training set. We then invoke the generalizing components stage, which clusters, deduplicates, and generalizes the learned components. This process produces a validated component repository containing 159 reusable components. For these steps, we use the Claude model. While Claude successfully solves all 30 scenarios, the first iteration produces 101 reusable components, and the second iteration increases this number to 159. We use the resulting component repository to seed the experiments on both the normal and challenge test sets, regardless of the model used on the test set.

\begin{wraptable}{r}{0.56\textwidth}
\vspace{-15pt}
\centering
\small
\setlength{\tabcolsep}{1pt}
\begin{tabular}{l@{\hskip 4pt}l@{\hskip 4pt}cc@{\hskip 6pt}cc}
\multirow{2}{*}{Method} & \multirow{2}{*}{Model} & \multicolumn{2}{c}{Normal} & \multicolumn{2}{c}{Challenge } \\
\cmidrule(lr){3-4} \cmidrule(lr){5-6}
& & TGC & SGC & TGC & SGC \\
\hline
Claude Code & Claude Opus 4.5 & 66.0 & 56.6 & -- & -- \\
Smolagents Code & Claude Opus 4.5 & 70.0 & 60.4 & -- & -- \\
OpenAI Solo & Claude Opus 4.5 & 68.0 & 56.6 & -- & -- \\
ReAct & Claude Opus 4.5 & 61.0 & 52.8 & -- & -- \\
ReAct Short & Claude Opus 4.5 & 64.0 & 54.7 & -- & -- \\
\hline
Alibaba AgentRL & Qwen3-14B & 86.9 & 80.4 & 67.6 & 50.4 \\
IBM CUGA & GPT-4.1 & 73.2 & 62.5 & 57.6 & 48.2 \\
LOOP & Qwen2.5-32B & 72.6 & 53.6 & 47.2 & 28.8 \\
\hline
\baselinemethodname (our baseline) & GPT-OSS 120B & 0.0 & 0.0 & 0.7 & 0.7 \\
\baselinemethodname (our baseline) & Claude Sonnet 4.6 & 96.4 & 96.4 & 83.2 & 82.0 \\
\hline
\methodname{} & GPT-OSS 120B & 62.5 & 62.5 & 41.0 & 41.0 \\
\methodname{} & Claude Sonnet 4.6 & {\bf 98.2} & {\bf 98.2} & {\bf 98.3} & {\bf 97.8} \\
\end{tabular}
\caption{
\label{tab:baseline-comparison}
Comparison of our scenario-level policy synthesis approach with task-level baselines on the AppWorld normal and challenge test sets. The challenge set includes applications (Gmail, Amazon) not present in the training set.
}
\vspace{-25pt}
\end{wraptable}

During test set evaluation, new components continue to be learned from successful policies and accumulated in the learned component repository. After processing a fixed number of debugging iterations (20 in our experiments), the generalization process is invoked again to consolidate newly learned components with the existing validated repository, further extending the set of reusable components available for subsequent scenarios.

\paragraph{Model Configurations.} We evaluate with two language models, accessed via API calls: \textbf{Claude Sonnet 4.6} (frontier model with strong code generation) and OpenAI \textbf{GPT OSS 120B} (open-source model at different capability level). We compare our \methodname{} configuration (with component retrieval, learning, and reuse) against a \baselinemethodname configuration (without these features) to isolate the contribution of dynamic component learning.

\paragraph{Additional baselines.} We include the top three methods from the AppWorld leaderboard: Alibaba Cloud ApsaraLab AgentRL using Qwen3-14B, IBM CUGA \cite{marreed-et-al-arxiv2025} using GPT-4.1, and LOOP \cite{chen-et-al-arxiv2025} using Qwen2.5-32B, as well as
agent architectures that use Claude Opus 4.5 from \citet{bandel-et-al-arxiv2026}.

Table~\ref{tab:baseline-comparison} summarizes results. The leaderboard methods and Bandel et al. methods operate at the task level rather than synthesizing scenario-level policies, resulting in lower SGC than TGC. In contrast, our dynamic policy-learning architecture, as well as our baseline, focuses on generalization across scenarios and improving efficiency through reusable components. The slight increase in TGC occurs when the policy computed in the final debugging iteration solves some of the tasks within a scenario.

\paragraph{Any-time Results.}

Rather than reporting only final task success, our evaluation emphasizes  how quickly success is achieved as a function of debugging effort and inference cost. This perspective is particularly appropriate for dynamic and interactive systems, where efficiency and reuse are central.

\begin{wrapfigure}{r}{0.48\textwidth}
    \vspace{-10pt}
    \centering
    \includegraphics[width=0.47\textwidth]{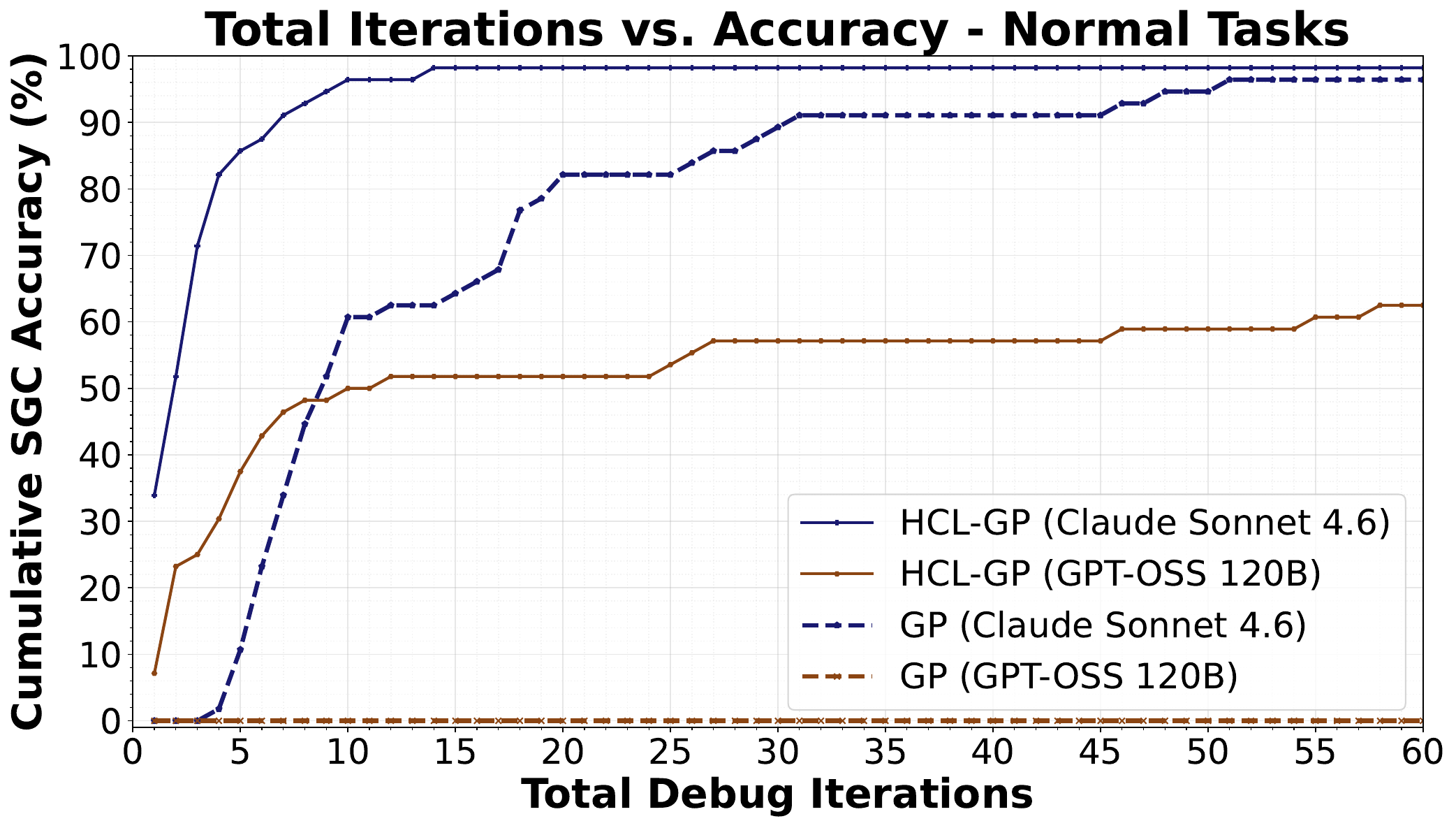}
    
    
    \includegraphics[width=0.47\textwidth]{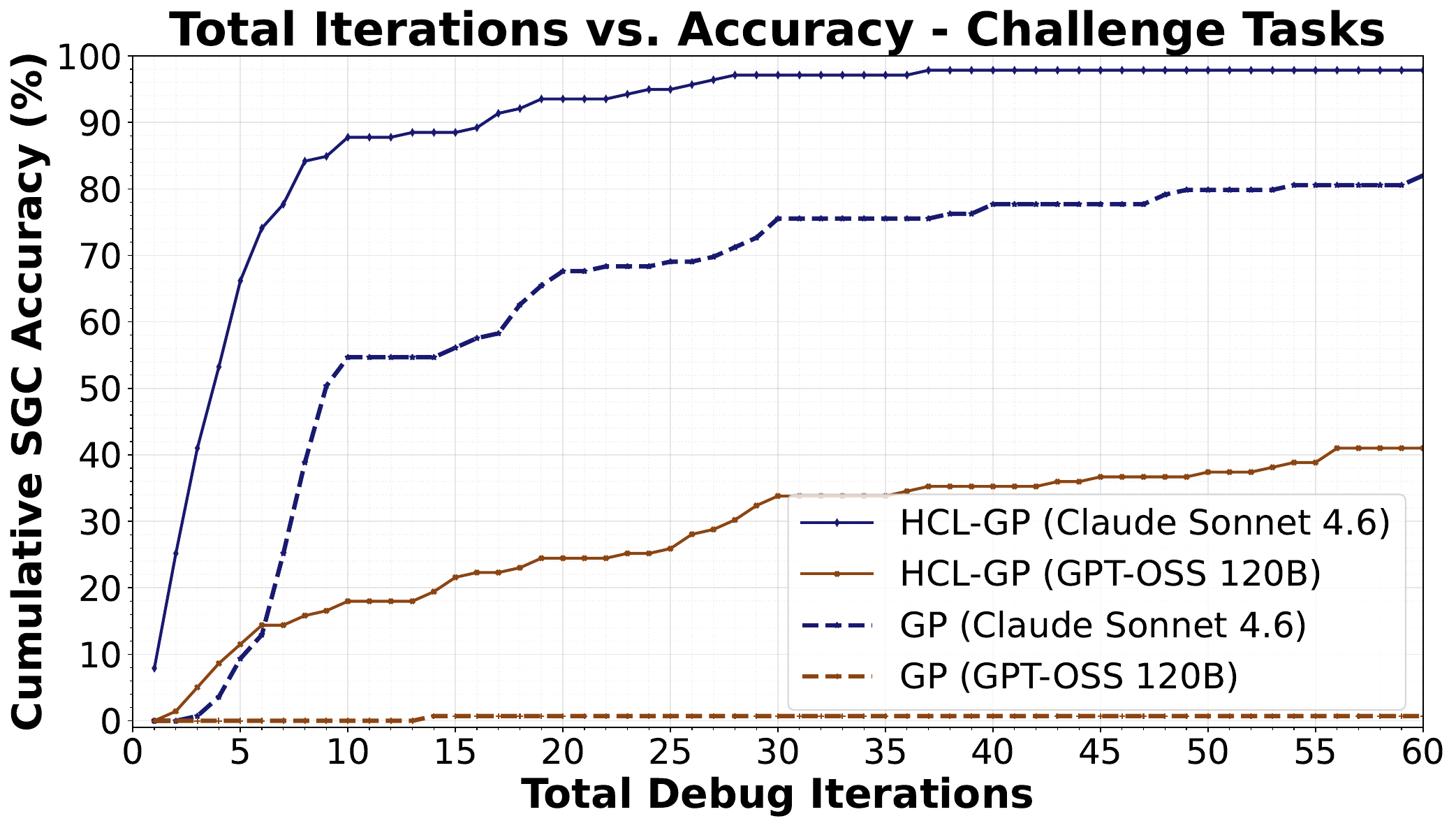}
    \caption{Cumulative SGC per total debugging iterations on the AppWorld normal (top) and challenge (bottom) test sets.}
    \label{fig:iters}
   \vspace{-5pt}
\end{wrapfigure}

\paragraph{Iteration Efficiency.}
Figure~\ref{fig:iters} shows cumulative SGC as a function of total debugging iterations on the normal and challenge test sets. We compare our proposed approach \methodname{} against the baseline variant \baselinemethodname for both Claude and GPT-OSS models.
As a reminder, after 20 and then 40 iterations \methodname{} performs generalization of freshly learned and previously existing components, and updates the set of validated components.

The results demonstrate that all methods benefit from additional debugging iterations, confirming that execution feedback and iterative repair are effective in this setting. However, the efficiency with which methods convert debugging effort into solved scenarios varies substantially. While Claude consistently outperforms GPT-OSS across all configurations, the largest performance gap appears between the \baselinemethodname and \methodname{} variants for GPT-OSS, where \baselinemethodname achieves near-zero success (0.0\% on Normal, 0.7\% on Challenge), demonstrating that without component reuse, the open-source model cannot effectively solve any of these scenarios.

For Claude, the \baselinemethodname variant achieves strong performance (96.4\% on Normal, 82.0\% on Challenge), though \methodname{} reaches even higher success rates (98.2\% and 97.8\% respectively) while requiring fewer debugging iterations. In contrast, \methodname{} consistently attains higher success earlier in the iteration budget, demonstrating that reusable components accelerate convergence. This advantage is particularly striking on the challenge set, where \methodname{} improves upon \baselinemethodname by 15.8 percentage points (82.0\% $\rightarrow$ 97.8\%), compared to only 1.8 points on the normal set. This substantial gap demonstrates that learned components provide critical support when encountering unfamiliar applications.

The \methodname{} approach dramatically improves GPT-OSS performance from near-zero to 62.5\% on Normal and 41.0\% on Challenge tasks. However, this still falls well short of Claude's performance, suggesting that while component reuse is essential for open-source models, base model capability remains a significant factor.

Overall, \methodname{} policies maintain a consistent advantage over their \baselinemethodname counterparts, solving more scenarios with fewer iterations. Notably, the challenge dataset includes applications such as Gmail and Amazon that do not appear in the training set. As a result, the initial component repository does not contain application-specific components for these applications. Despite this, \methodname{} benefits from previously learned generalized components that transfer across applications, contributing to improved performance on unseen domains.

\begin{wrapfigure}{r}{0.48\textwidth}
    \centering
    \includegraphics[width=0.47\textwidth]{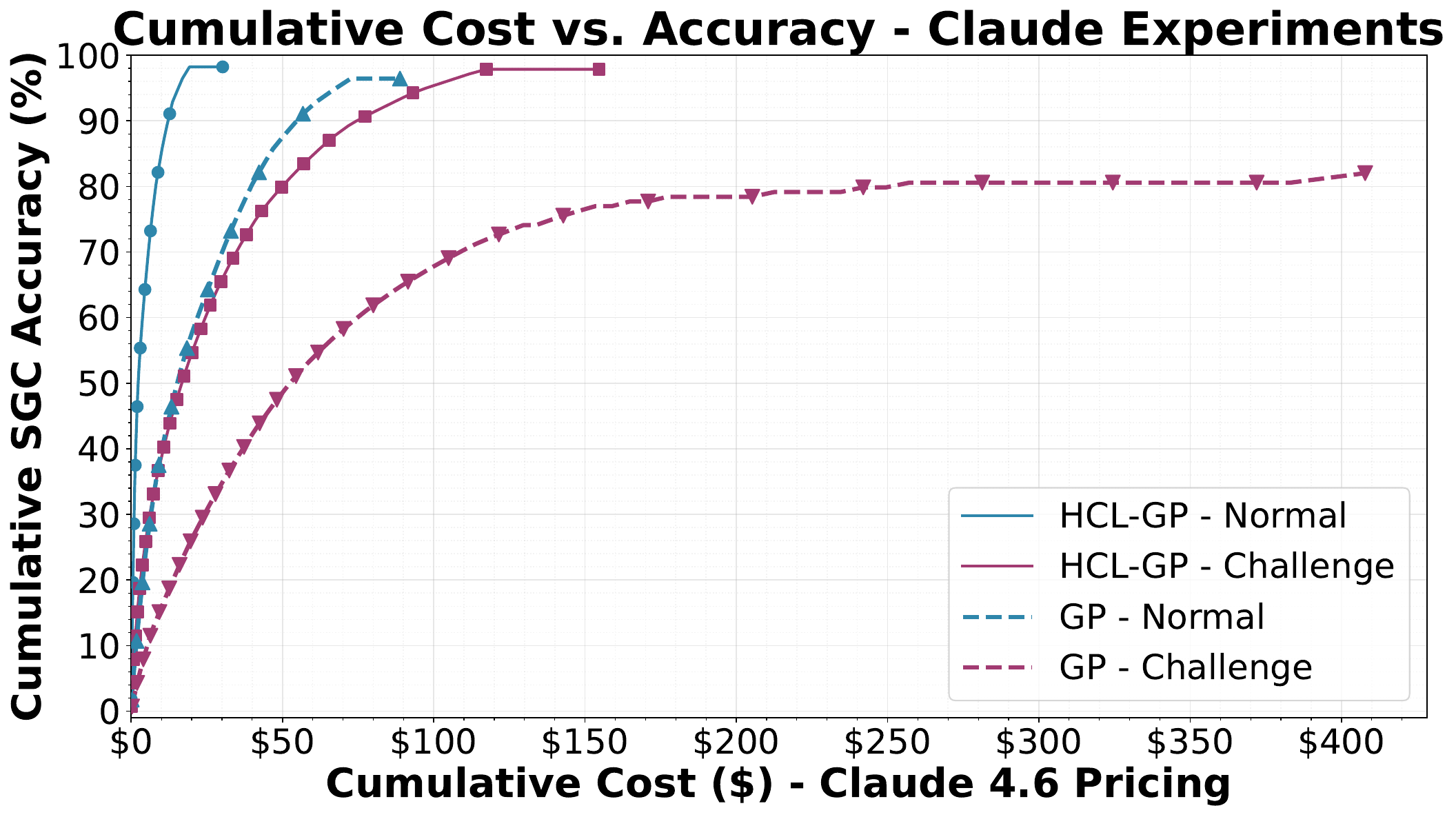}
    \caption{Cumulative SGC as a function of 
    inference cost for using Claude~4.6 pricing.}
    \label{fig:cost-accuracy}
    \vspace{-5pt}
\end{wrapfigure}

\paragraph{Cost-Accuracy Trade-offs.}

For cost analysis, Figure~\ref{fig:cost-accuracy} reports cumulative SGC as a function of cumulative inference cost using Claude Sonnet 4.6 pricing ($\$3$/$\$15$ per million input/output tokens).

Across all settings, success increases rapidly at low cost and exhibits diminishing returns as cost grows. Dynamic reuse shifts this trade-off favorably: higher success is achieved earlier in the cost budget, particularly on challenge tasks. This result reinforces the interpretation from the iteration-based analysis. Dynamic reuse does not merely improve final success, but more efficiently converts inference budget into solved tasks by reducing redundant policy synthesis and debugging.

The cost analysis is particularly relevant when comparing against task-level methods. While we do not have detailed cost breakdowns for all baseline methods, the fundamental difference in approach suggests that scenario-level policy synthesis with reuse should be more cost-effective: rather than solving each task independently (often repeating similar reasoning across tasks), our approach synthesizes a single parameterized policy per scenario and reuses learned components across scenarios.

\paragraph{Summary of Key Findings.}
Taken together, these results highlight three key observations:

\noindent\textbf{1. Scenario-level synthesis is effective:} Even without dynamic reuse, synthesizing parameterized policies at the scenario level can substantially outperform task-level methods in terms of SGC, as demonstrated by Claude's 96.4\% baseline. However, this advantage depends on sufficient base model capability, as GPT-OSS's baseline performance (0.0\%) falls below task-level methods.

\noindent\textbf{2. Dynamic reuse improves efficiency:} The advantages of reusable components are most pronounced in terms of iteration efficiency and cost-accuracy trade-offs. Dynamic reuse enables the system to achieve high success rates earlier in the debugging budget, particularly on challenging scenarios.

\noindent\textbf{3. Components transfer across applications:} Learned components generalize beyond the specific applications they were extracted from, as evidenced by improved performance on challenge scenarios that include previously unseen applications (Gmail, Amazon).

These findings support the view that reusable, compositional structure, learned dynamically from prior executions, plays a critical role in scaling interactive agents to complex, multi-step domains. The combination of scenario-level policy synthesis and dynamic component learning provides both higher final accuracy and more favorable efficiency characteristics compared to task-level approaches.

\paragraph{Component Usage Analysis.}

Table~\ref{tab:component-usage} summarizes component usage patterns across test sets. Components are categorized as Seed unchanged (S), Seed modified (M), and Learned (L).

\begin{wraptable}{r}{0.55\textwidth}
\vspace{-10pt}
\centering
\small
\setlength{\tabcolsep}{1pt}
\begin{tabular}{l@{\hskip 3pt}cc}
\toprule
\textbf{Metric} & \textbf{Normal (S/M/L)} & \textbf{Challenge (S/M/L)} \\
\midrule
Total Used
& 42 / 1 / 203 & 46 / 2 / 638 \\
Utilization Rate \% & 26 / 20 / 75 & 29 / 40 / 60 \\
Per Scenario & 2.9 / 0.8 / 3.8 & 1.5 / 0.9 / 6.4 \\
Reuse Rate & 3.7 / 46.0 / 1.0 & 4.5 / 64.5 / 1.4 \\
Multi-use \% & 59.5 / 100 / 1.5 & 58.7 / 100 / 11.1 \\
\bottomrule
\end{tabular}
\caption{Component direct usage statistics. \textbf{Total}: unique components used. \textbf{Utilization Rate}: \% of available used. \textbf{Components per Scenario}: mean per scenario. \textbf{Reuse Rate}: mean scenarios per components. \textbf{Multi-use}: \% used in 2+ scenarios.}
\label{tab:component-usage}
\end{wraptable}

Three key patterns emerge. First, learned components show high utilization (60--75\%) compared to seed components (26--29\%), indicating the system learns task-relevant components. Second, the two modified components are used directly or indirectly (through other components) across scenarios, with \texttt{get\_supervisor\_credentials} used in all scenarios. Seed components provide stable reuse, while learned components are mostly task-specific. Third, Challenge scenarios require more learned components per scenario (6.4 vs 3.8), and the 3$\times$ growth in unique learned components (203$\rightarrow$638) exceeds the 2.5$\times$ dataset size increase (56$\rightarrow$139 scenarios), indicating more diverse component learning on harder tasks.

\section{Challenges and Discussions}

The results highlight several insights about dynamic reuse and policy generalization in structured interactive environments. First, reusable components' benefits are strongly task-dependent. On simpler scenarios, execution feedback and iterative debugging is often sufficient, with reuse primarily improving efficiency by reducing iterations.
In contrast, on more challenging scenarios, reusable components play a more substantive role, providing structural guidance that helps the agent avoid repeated errors and recover from failures more effectively. This suggests that dynamic reuse is particularly valuable in domains characterized by longer execution horizons or more brittle workflows.

Second, components learned from one set of scenarios can transfer to unseen applications, as shown by improved performance on challenge scenarios with previously unseen applications. This indicates that the learned components capture procedural abstractions that extend beyond app-specific logic. At the same time, this form of generalization is opportunistic rather than guaranteed: it emerges through execution grounding and consolidation rather than through explicit domain modeling. Understanding the conditions under which such transfer is reliable remains an open research question.

The architecture also exposes several challenges that become increasingly important when moving beyond AppWorld. One challenge concerns assumptions about the availability of a well-defined \emph{meta-domain}. 
AppWorld provides fixed action spaces, execution semantics, and validation mechanisms. More general settings require dynamic discovery of actions, constraints, and validation signals.

A related challenge arises from the clean scenario structure provided by AppWorld. Scenarios group structurally similar tasks, enabling straightforward learning of scenario-level policies. In many real-world settings, however, such scenario boundaries may not be given a priori. Identifying which tasks belong to the same domain or which tasks are sufficiently similar to benefit from shared abstractions becomes nontrivial. This would require clustering based on inferred procedural structure, execution traces, or learned representations, rather than relying on externally provided scenario labels.

Scalability of reusable component evaluation presents another challenge. Although clustering coarsely groups similar components, generalization and deduplication still rely on LLM  reasoning and are sensitive to prompt length and context limits. As the number of components grows, their token footprint becomes increasingly important. More structured multi-stage evaluation pipelines, or incremental abstraction mechanisms, may be needed to sustain long-term learning.

\section{Summary and Future Work}

We presented a dynamic policy-learning architecture for structured interactive environments that integrates ideas from generalized planning and hierarchical decomposition with LLM-based agents. The approach operates at two levels: a per-scenario flow that synthesizes parameterized scenario-level policies, and an attempt-level process that extracts, evaluates, and generalizes reusable executable components. By separating execution, learning, and generalization, the system incrementally constructs a reusable component repository that supports compositional policy generation across scenarios.

Empirical results on the AppWorld benchmark demonstrate that dynamic reuse improves both iteration efficiency and cost--accuracy trade-offs, particularly on challenging scenarios that require longer and more structured decision-making. The results also highlight that reusable components need not be tied to specific applications: generalized components learned from one set of scenarios can transfer to previously unseen ones, supporting broader generalization within a shared meta-domain. 

There are several promising directions for future work. First, while reusable components are currently treated as flat executable units, richer structural organization could be explored, such as learning explicit dependency graphs or multi-level abstractions over components. Second, the current approach relies primarily on semantic similarity for clustering, with generalization and deduplication combined into a single LLM invocation. While clustering itself could be further improved, separating the generalization and deduplication stages would substantially reduce the token footprint of each LLM call. In practice, we have observed that the combined operation can exceed the maximum context length of the underlying model, making this separation an important direction for improving scalability. Finally, applying the architecture to domains beyond AppWorld would help assess the broader applicability and impact of the proposed approach.

\bibliographystyle{plainnat}

\newpage

\appendix

\appendix
\section*{Appendix}
\addcontentsline{toc}{section}{Appendix}

\begin{wraptable}{r}{0.55\textwidth}
\vspace{-10pt}
\centering
\small
\setlength{\tabcolsep}{2pt}
\begin{tabular}{l@{\hskip 4pt}cccccc}

Configuration & MRR & MAP & R@5 & R@10 & R@20 \\
\hline
OpenSearch + BAAI & 0.24 & 0.43 & 35.2 & 51.4 & 70.2 \\
FAISS + BAAI      & 0.24 & 0.44 & 36.2 & 56.5 & 68.3 \\
FAISS + Nomic     & 0.21 & 0.35 & 29.0 & 47.7 & 63.8 \\

\end{tabular}
\caption{API retrieval performance on the AppWorld training set for different embedding and retrieval configurations.
Metrics include mean reciprocal rank (MRR), mean average precision (MAP), and recall at $k$ (R@5, R@10, R@20).}
\label{tab:search-ablation}
\end{wraptable}

\section{Search Effectiveness.}

In this ablation study, we evaluate the \emph{Search} component on the 30 training scenarios under different embedding models and retrieval configurations. We evaluate both FAISS-based approximate nearest-neighbor search \cite{johnson-et-al-tbd2021} and an OpenSearch-based retrieval backend \url{https://opensearch.org/} for semantic search. Note that Nomic is short for the \textit{nomic-ai/nomic-embed-text-v1.5} embedding model.
For the training set, AppWorld provides ground-truth annotations specifying the APIs required to solve each task.
Using this information, we evaluate the Search component in isolation by measuring its ability to retrieve relevant APIs.

Table~\ref{tab:search-ablation} reports standard information retrieval metrics: mean reciprocal rank (MRR), mean average precision (MAP), recall at 5 (R@5), recall at 10 (R@10), and recall at 20 (R@20).
Recall values indicate the fraction of tasks for which at least one correct API appears within the top-$k$ retrieved results.
Based on these results, we select a FAISS-based configuration using the BAAI embedding model for both semantic search and reusable component clustering in all experiments.

Table~\ref{tab:search-ablation} shows that FAISS-based retrieval with BAAI achieves the strongest recall at lower cutoffs (R@5 and R@10), while maintaining competitive MRR and MAP.

\section{Agent Prompt Specifications}
\label{sec:agent-prompts}

This appendix documents the prompts used for the Policy Generator Agent, Listing 1 and 2, Task Abstraction and Parameterization Agent, Listing 3, Decomposition Agent, Listing 4, and Generalize and Deduplicate Agent, Listing 5 and 6.  

\begin{listing*}[h]
\scriptsize
\begin{minted}{text}
# AI Coding Agent for AppWorld Policy Generation

You are an AI coding agent specializing in generating generalized policy in Python code. Your goal is to 
design a  **generalized policy (in Python code)** for each scenario (collection of similar tasks) that 
can solve any of the tasks under that scenario. The tasks are all from the Appworld benchmark 
(https://appworld.dev/task-explorer).

Each policy must:
* Handle all possible points of failure (e.g., insufficient balance, expired cards, authentication issues).
* Be modular — reusable components separated from scenario-specific logic.
* Be self-contained, robust, and executable within AppWorld.
* Assume that all tasks are solvable, though they may include realistic hurdles or distractors.

## Resources to Use
### 1. All APIs in AppWorld
To get a list of apps:
```python
print(apis.api_docs.show_app_descriptions())
```

Output:
```
["amazon : An online shopping platform where users can buy a wide variety of products.",
 "spotify : A music streaming platform that provides access to a vast library of songs, albums, and 
 playlists.",...]
```
### 2. Available Reusable Components
The reusable components are organized into the following categories. **The full implementation code for 
these functions is provided separately - do not rewrite them, use them exactly as provided.**

### Category Overview:
1. **API Discovery** - Find and understand unknown APIs
2. **Learned API Utilities** - Pre-discovered components you can call directly - You will be provided 
with the most relevant ones for your task to start with. You can implement new components based 
on the task at hand.
3. **Recommended APIs** - Recommended APIs relevant to your task.

# CATEGORY 1: API Discovery 
You can discover APIs using:
```python
# Search for APIs
search_results = apis.api_docs.search_api_docs(
    query="your search terms",
    page_limit=20)

# Show descriptions for all APIs available for a given app
api_descriptions = apis.api_docs.show_api_descriptions(
    app_name="app_name")

# Get API structure
api_info = apis.api_docs.show_api_doc(
    app_name="app_name",
    api_name="api_name")
```
**Important Notes on API Discovery:**
- Do NOT guess API names like `list_contacts`, `get_contacts`, `find_contacts`, `get_supervisor_info`, 
`list_accounts`, `get_user_account_info` without checking first - these don't exist
- Always use `search_api_docs()` or `show_api_descriptions()` to discover available APIs
- Check the API structure with `show_api_doc()` before calling it as it may require additional 
required fields, such as access_token
- Different apps may have different parameter names (e.g., `relationship` vs `contact_type`)
- For apps that require pagination (e.g., `page_limit`), handle them in your code so you consider all 
results. 

## CATEGORY 2: Learned API Utilities (Includes Dynamically Loaded Suggestions for This Scenario - Try 
to use them to your maximal extent, and implement new components that you deem fit for the task at 
hand.) The following pre-filtered utilities are ranked from highest to lowest relevance for your scenario.

{{SHORTLISTED_UTILITIES}}

## CATEGORY 3: Recommended APIs
The following are ranked APIs from highest to lowest relevance to your task.

**Important:** These are the most relevant APIs for your specific task. Start by exploring these APIs 
first before searching for alternatives.
**When to use**: For operations not covered by learned utilities but semantically relevant to your task. 
The top-ranked APIs are most relevant to accomplishing the task.

{{SUGGESTED_APPWORLD_APIS}}
\end{minted}
\caption{Policy Generation Agent prompt used to synthesize scenario-level policies.}
\label{listing:policygen-prompt}
\end{listing*}

\begin{listing*}[h]
\scriptsize
\begin{minted}{text}
**How to use**:
Call directly if you're confident 
   - `apis.spotify.remove_song_from_library(song_id=X, access_token=Y)`
   - But you must handle errors yourself

## Requirements
### 1. Solvability. All tasks are designed to be solvable. Consider all contingencies for API call outputs.
### 2. Number Answers. Convert answers to nearest integer. Answer format: `answer=10` not "ten".
### 3. Code Style
* Maintain syntactically valid, readable Python code
* Use `apis.<app>.<endpoint>()` if calling an end point directly
* Integrate API calls with control flow
### 4. Error Handling. While reusable components handle many errors internally, your scenario-specific 
code should:
* Check for empty list returns from reusable functions
* Check for `None` returns when fetching data
* Handle exceptions when calling APIs that aren't wrapped in reusable components
* Implement appropriate logic (retry, skip, or terminate) based on requirements
### 5. Progress Tracking & Debugging. Include `print()` statements throughout the code:
* Use clear step dividers: `print("="*80)`
* Use status markers: `[OK]`, `[FAIL]`, `[WARNING]`
* Print variable states after important operations
### 6. File System Reference. Use file_system app APIs, never `os` or `shutil`.
### 7. Current Date and Time. Use `datetime.now()` and `get_date_range()`. Never rely on your existing 
knowledge of what the current date is.
### 8. Data Selection
* References to people mean phone contacts
* Get supervisor info from `get_supervisor_credentials()`
* Filter results carefully before selecting
* Do not make assumptions - always verify
### 9. Task-Specific Terminology. Follow exactly the terms, labels, and tags specified in the task.
### 10. Contact Matching Safety. Prefer email-first matching, then exact full-name match only if unique.
### 11. Task Completion
* Call `apis.supervisor.complete_task()` when done
* If task asks for information, pass it as: `apis.supervisor.complete_task(answer=<answer>)`
* Answer should be a valid number, integer, or string only
* You can pass `status="fail"` if you cannot solve the task
### 12. Autonomous Decision Making. Make all decisions autonomously. No clarifications.
### 13. Learned Utilities vs Discovery Pattern
**Use Learned Utilities (Preferred):**
* If a learned component covers your use case, use it instead of rediscovering APIs.
**Discover When Needed**
* For new operations not covered by learned components, use API discovery as shown above.

## Output Structure
For each scenario, provide:
1. **Assume all reusable component implementations will be added to your code** (provided separately)
2. **Create scenario-specific helper functions** as needed
3. **Follow the proven structure** with clear step markers (STEP 1, STEP 2, etc.)
4. **Use learned utilities when available**
5. **Use discovery pattern for unknown operations**
6. **Include the main policy function** that orchestrates all the steps

**Structure your code like this:**

```python
def scenario_X_policy(param1, param2, ...):
    Generalized policy for scenario X.
    Args:
        param1: Description of parameter 1
        param2: Description of parameter 2
    print("="*80)
    print("SCENARIO X: [Description]")
    print("="*80)
    
    # STEP 1: Discover APIs
    print("\n[STEP 1] Discovering APIs...")
    
    # Use search_api_docs to find needed APIs
    search_results = apis.api_docs.search_api_docs(query="search terms", page_limit=20)
    
    # STEP 2: Get API structures
    print("\n[STEP 2] Getting API structures...")
    
    api_info = apis.api_docs.show_api_doc(app_name="app_name",api_name="api_name")
    # STEP 3: Authentication
    print("\\n[STEP 3] Authenticating...")
 ...
\end{minted}
\caption{Policy Generation Agent prompt (continued)}
\label{listing:policygen-prompt-cont}
\end{listing*}

\begin{listing*}[h]
\scriptsize
\begin{minted}{text}
#You are defining the callable Python function signature for a scenario policy. Input is a 
SCENARIO DESCRIPTION containing multiple tasks from the same scenario. Your job:

1) Propose a minimal, stable list of policy function parameters that the runner can pass at call time.
2) Every policy MUST accept `task_text: str` (required).
3) Do NOT include parameters that cannot be derived from task_text or discovered via AppWorld APIs at 
runtime.
4) Keep params general across tasks in the scenario (avoid overfitting to one task).
5) IMPORTANT: Based on each of the 3 tasks mentioned in the scenario description, assign param_values 
to each parameter you define, except the task_text parameter. This is mandatory.
6) IMPORTANT: Identify which AppWorld apps are needed for this scenario.
Available Apps[
 "amazon : An online shopping platform where users can buy a wide variety of products.",
 "spotify : A music streaming platform that provides access to a vast library of songs, albums, and 
 playlists.",...] Add an "apps_used" field to your JSON output listing the apps mentioned or implied 
 in the tasks.
7) IMPORTANT: Plan out the high-level steps needed to accomplish the task. What are the steps you will 
need take to complete the task?
   - For each step, be as specific and detailed as possible. Include all the steps that are necessary 
   including logging in to an app or getting the necessary credentials. 
   - ALWAYS explicitly mention which app(s) will be used in that step (e.g., "Use file_system to 
   read ...", "Use venmo to send...")
   - If a step involves multiple apps, try to break this step into multiple steps, rather than having a 
   step that mentions more than one app
   - Use the exact app names from the list above (e.g., "file_system" not "filesystem")
   Add a "steps_needed" field to your JSON output listing all the steps needed to accomplish the task.

Output MUST be valid JSON with schema: 

### Example (for guidance only)
Scenario question:
"What is the title of the most-liked song in my Spotify playlists."
A good parameter interface for this scenario would be:
{{"apps_used": ["spotify", "supervisor"],
  "steps_needed": [
    "Get the credentials needed from Supervisor for login",
    "Login to Spotify",
    "Query Spotify for the user's playlists",
    "Find the most-liked song in the remaining playlists",
    "Return the title of the most-liked song"],
  "params": [
    {{"name": "task_text",
      "type": "str",
      "description": "Full natural-language instruction for the current task instance",
      "required": true,
      "default": null,
      "source": "scenario_text",
      "extraction_hint": "Use as-is",
      "param_values": {{}}}},
    {{"name": "include_private_playlists",
      "type": "bool",
      "description": "Whether to include the user's private playlists when interpreting 
      'my Spotify playlists'",
      "required": false,
      "default": true,
      "source": "scenario_text",
      "extraction_hint": "If scenario_text explicitly excludes private playlists, set false; otherwise 
      default true",
      "param_values": {{"task_1": True, "task_2": False, "task_3": True}}}}]}}

Notes:
- These parameters express TASK INTENT, not API details.
- Optional parameters must always have safe defaults.
- The policy must still work if optional parameters are null.

Allowed types: str, int, float, bool, list[str]
Allowed sources: task_texts, constant, derived
Return JSON only (no extra text).

SCENARIO DESCRIPTION: {scenario_description}
\end{minted}
\caption{Task Abstraction and 
Parameterization Agent prompt used to infer scenario-level parameters and high-level task structure from natural language descriptions.}
\label{listing:param-prompt}
\end{listing*}

\begin{listing*}[h]
\scriptsize
\begin{minted}{text}
#You are a code extraction agent. Your task is to extract reusable components from a successful 
policy implementation.

The policy has successfully solved a task. Now extract any reusable components:

**What to Extract:**
- API calls that were discovered and successfully used (wrap them in utility functions)
- Data transformations, validations, calculations
- Filtering patterns, field extraction logic
- Any repeated code blocks or inline logic that could be reused

**Make components general-purpose, not scenario-specific.**

## CRITICAL: Output Format

You MUST provide EXACTLY two sections in this format:

### REUSABLE COMPONENTS
```python
# Component 1 - <Description> (Discovered from Scenario <ID>)

def component_name(params):
    \"\"\"
    Clear docstring explaining what this does.
    
    Args:
        param1: Description
        param2: Description
    
    Returns:
        Description of return value
    \"\"\"
    # Implementation

# Component 2 - <Description> (Discovered from Scenario <ID>)

def another_component(params):
    \"\"\"
    Clear docstring.
    \"\"\"
    # Implementation
```

### UPDATED POLICY
```python
def scenario_<id>_policy(params):
    \"\"\"
    Updated policy using the extracted reusable components.
    \"\"\"
    # Implementation using the new components
```

## Rules:
- Use EXACTLY these two headings
- Each section must have ONE ```python code block
- NO text outside these two sections
- NO explanations or commentary
\end{minted}
\caption{Decomposition Agent prompt used to identify and factor reusable components from successfully executed policies.}
\label{listing:learn-utility-prompt}
\end{listing*}

\begin{listing*}[h]
\scriptsize
\begin{minted}{text}
#You are a code generalization expert specializing in creating reusable, generic components from 
specific implementations. Your task has TWO parts:

## Part 1: Generalize Multi-Function Clusters
Analyze clusters of similar functions learned from different scenarios and create ONE generalized 
function per cluster that:
1. Works for all functions in the cluster
2. Has a generic name (e.g., 'authenticate' instead of 'authenticate_venmo')
3. Uses parameters to handle app-specific or scenario-specific differences
4. Includes comprehensive docstring explaining the generalization
5. Maintains all functionality from the original functions

**IMPORTANT - Clustering Flexibility:**
- The initial clustering may not be perfect. You have the flexibility to:
  - **Split a cluster** if functions are too different and shouldn't be merged (keep them as separate 
  functions)
  - **Merge across clusters** if you notice functions from different clusters that should actually be 
  combined
  - **Include helper functions** from other clusters if they would improve the generalized implementation
- Use your judgment: if functions in a cluster are fundamentally different despite similarity scores, 
keep them separate and document why in the docstring
- Conversely, if you see opportunities to combine functions across clusters for better reusability, do so

## Part 2: Update Dependent Single Functions (if provided)
Some single functions may call the original functions from Part 1. After generalizing those functions, 
you MUST:
1. Update these single functions to call your NEW generalized functions
2. Adjust parameters to match the new generalized function signatures
3. Preserve all other functionality in these single functions

## Key Principles:
**Generalization:**
- Remove app-specific names when possible (venmo_login → login, with app parameter). However, if the
function is actually only meant for a particular app, then keep the app name in the name of the function 
and include the app name in the docstring.
- Use parameters for variations instead of separate functions
- If the function have different paramaters and different return values but perform similar function, 
still merge them.
- **IMPORTANT**: You can call other generalized functions you create in your implementations - build 
on top of simpler functions
- **IMPORTANT**: DO NOT call the original functions (e.g., original_function_1, original_function_2) - 
they will not be available. Only use the generalized functions you create and the AppWorld APIs (apis.*)
**Smart Parameter Design:**
- You do NOT need to accommodate every possible parameter variation from all versions
- Choose the MOST SENSIBLE parameter set that covers the core functionality
- Add an app parameter (e.g., app_name="spotify") to make functions work across different apps
- Don't create overly complex signatures trying to handle every edge case
- Focus on the common use case and let the app parameter handle app-specific differences
**Updating Dependent Functions:**
- When updating single functions that call original functions, replace those calls with your new 
generalized functions
- Adjust parameters carefully - the generalized function may have different signatures
- Example: If `original_login_spotify()` becomes `login(app_name='spotify')`, update all calls 
accordingly
- Preserve all other logic in the single function
**Deduplication:**
- If two functions do exactly the same thing, call the same APIs, keep only one
- If functions are similar but have minor differences, merge them with parameters
**Preservation:**
- Keep all unique functionality. If you believe that they should not be clustered together, then keep 
both functions and indicate that in the docstring
- Don't lose edge case handling
- Maintain error handling and validation
**Code Reuse:**
- You can and SHOULD call other generalized functions you've created to build more complex functionality
- Example: If you create a generic `login()` function, you can use it in a `get_authenticated_user()` 
function
- This creates a hierarchy of reusable components
- Only call: (1) Your own generalized functions, (2) AppWorld APIs (apis.*)
- NEVER call the original learned functions - they won't exist in the final code

**Code Quality:**
- Clear, descriptive function names
- Comprehensive docstrings
- Type hints where appropriate
- Well-structured, readable code

\end{minted}
\caption{Generalize and Deduplicate Agent prompt used to deduplicate, generalize, and update learned component repository across scenarios.}
\label{listing:evaluaterc-prompt}
\end{listing*}

\begin{listing*}[h]
\scriptsize
\begin{minted}{text}
## Output Format:
You MUST provide your output in this exact format:

# GENERALIZED COMPONENTS
# Generated from multiple learned scenarios

def generalized_function_name(param1, param2, app_name=None):
    Brief description of what this function does.
    
    Generalized from:
    - original_function_1 (Scenario XXX)
    - original_function_2 (Scenario YYY)
    
    Args:
        param1: Description
        param2: Description
        app_name: App name (e.g., 'spotify', 'phone') to determine which API to use
    
    Returns:
        Description of return value
    
    Example:
        # Login to Spotify
        result = generalized_function_name(param1, param2, app_name='spotify')
        
        # Login to Phone
        result = generalized_function_name(param1, param2, app_name='phone')
    # Implementation using app_name to call appropriate API
    if app_name == 'spotify':
        return apis.spotify.some_method(param1, param2)
    elif app_name == 'phone':
        return apis.phone.some_method(param1, param2)
    # etc.

def higher_level_function(param1, app_name):
    Example of calling other generalized functions.
    
    This demonstrates building on top of simpler generalized functions.
    DO NOT call original functions - they won't be available!
    # CORRECT: Call other generalized functions you created
    result = generalized_function_name(param1, "default", app_name=app_name)
    
    # CORRECT: Call AppWorld APIs directly
    user_data = apis.supervisor.get_user()
    
    # WRONG: Do NOT call original functions like original_function_1()
    # They will not be available in the final code!
    return result

# If Part 2 was provided, include updated single functions here
def updated_single_function(param):
    Single function that was updated to use generalized functions.
    
    Originally from: scenario_xyz/learned_function.py
    Updated to call: generalized_function_name() instead of original_function_1()
    # CORRECT: Call the new generalized function with proper parameters
    result = generalized_function_name(param, "value", app_name='spotify')
    
    # Rest of the original logic preserved
    return result

# Add more generalized and updated functions as needed

## Rules:
- Output ONLY Python code in a single code block
- Include ALL functions: generalized (Part 1) AND updated single functions (Part 2, if provided)
- Start with comment: # GENERALIZED COMPONENTS
- Each function must have a docstring listing original functions it was generalized from
- For updated single functions, note in docstring what was changed
- NO explanations outside the code block
- NO markdown except the code fence

\end{minted}
\caption{Generalize and Deduplicate Agent prompt (continued)}
\label{listing:evaluaterc-prompt}
\end{listing*}

\end{document}